\documentclass{article}
\usepackage{spconf,amsmath,epsfig,subfig}
\usepackage{color}
\usepackage{graphicx}
\usepackage{multirow}
\usepackage{subfig}

\begin{document}\sloppy

% Example definitions.
% --------------------
\def\x{{\mathbf x}}
\def\L{{\cal L}}

% Title.
% ------
\title{Select-Additive Learning: \\
Improving Generalization in Multimodal Sentiment Analysis}
%
% Single address.
% ---------------
\name{Haohan Wang, Aaksha Meghawat, Louis-Philippe Morency \textnormal{and} Eric P. Xing}
\address{Language Technologies Institute\\
School of Computer Science\\
Carnegie Mellon University \\
\textit{\{haohanw, aaksham, morency, epxing\}@cs.cmu.edu}
}

\maketitle

\begin{abstract}
Multimodal sentiment analysis is drawing an increasing amount of attention these days. It enables mining of opinions in video reviews which are now available aplenty on online platforms. However, multimodal sentiment analysis has only a few high-quality data sets annotated for training machine learning algorithms. These limited resources restrict the generalizability of models, where, for example, the unique characteristics of a few speakers (e.g., wearing glasses) may become a confounding factor for the sentiment classification task. In this paper, we propose a Select-Additive Learning (SAL) procedure that improves the generalizability of trained neural networks for multimodal sentiment analysis.
In our experiments, we show that our SAL approach improves prediction accuracy significantly in all three modalities (verbal, acoustic, visual), as well as in their fusion. Our results show that SAL, even when trained on one dataset, achieves good generalization across two new test datasets.  
\end{abstract}
\section{Introduction}
Sentiment analysis is the automatic identification of the private state of a human mind with a focus on determining whether this state is positive, negative or neutral \cite{morency2011towards}. It has been extensively studied in the last few decades \cite{pang2008opinion}, primarily based on textual data. With the recent proliferation of online avenues for sharing multimedia content, people are posting more and more videos with opinions. The opinions are expressed through the spoken word (verbal modality), how these words are spoken (acoustic modality) and what gestures and facial expressions accompany the spoken words (visual modality). Multimodal sentiment analysis extends traditional textual sentiment analysis by analyzing all three modalities present in online videos, including acoustic and visual modalities \cite{kumar2012sentiment,wollmer2013youtube}.  

\begin{figure}[ht]
\subfloat[An illustrative data set]{\includegraphics[width=0.5\columnwidth]{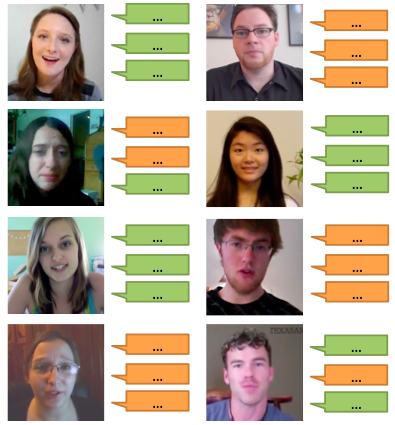}}
\hfill
\subfloat[Rules learned]{\includegraphics[width=0.4\columnwidth]{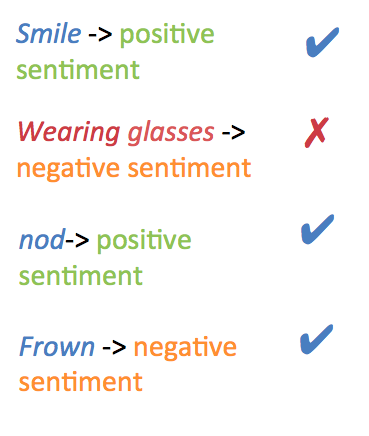}}
\hfill
\caption{An illustrative data set demonstrating the ``wearing glass'' as a confounding factor. Due to the limited amount of data, the model learns that wearing glasses means negative sentiment, which is only applicable to this training data set. (Orange denotes negative sentiment; green denotes positive sentiment; blue denotes correct rules \& red denotes incorrect rules).}
\label{fig:problem}
\end{figure}

To foster research in this area, a few datasets have been created with quality annotations for sentiment \cite{morency2011towards,rosas2013multimodal,zadeh2015micro}, but unfortunately the total number of annotations is still in the order of thousand samples. These limited-size resources make it challenging for conventional machine learning algorithms to generalize well across datasets. In these limited data scenarios, a unique characteristic of a few speakers in the training dataset (e.g., wearing glasses) can end up creating a confounding effect with the sentiment classification task.
% In addition to individual style of expression, a machine learning algorithm may also learn to recognize sentiment based on speaker's identity, in contrast to relevant features (e.g. facial expression) if the each individual tends to express only one polarity of sentiment.
Fig.~\ref{fig:problem}(a) shows one illustrative example where limited data can bring in learning and generalization challenges. Since in this example all individuals with glasses happen to be expressing negative sentiments, the classifier ends up learning an association between visual appearance of wearing glasses and negative sentiment (see Fig.~\ref{fig:problem}(b)) . 
%Fig.~\ref{fig:problem}(b) is an example set of decision rules a classifier can learn from this illustrative data set\footnote{Figure 1 illustrates "rules" that a classifier could learn on this dataset. In our paper, the proposed approach is based on neural networks where the confounding factor are encoded in the neural representations.}. In addition to a few authentic rules that are helpful for sentiment analysis (labeled with ``check mark''), it also learns a short-cut rule that is only applicable to this dataset (labeled with ``cross''). These short-cut rules are learned due to the dependency between the sentiment expressed and an individual's identity. 

The role that the visual appearance plays here is statistically known as a confounding factor \cite{ewers2006confounding,wang2016multiple}. 
To generalize across datasets and individuals, a robust multimodal sentiment classifier should not include features from a confounding factor. In other words, the prediction of the sentiment polarity should not be dependent on the speaker's unique characteristic, namely, identity of the speaker.

Before going further in this research agenda, we studied if the confounding factor also exists in the real-world multimodal sentiment analysis datasets. In the MOSI multimodal sentiment analysis data set \cite{zadeh2015micro}, we tested for the null hypothesis that sentiment is independent of an individual's identity. Chi-square test obtains a p-value of $1.202\times 10^{-19}$, which strongly suggests the dependence between individual identities and the expressed sentiment. Consequently, naively applying machine learning algorithms on this dataset will most likely result in a suboptimal model that misinterprets an individual's identity as prescient information for sentiment.

%Two remarks need to be made to clarify potential misunderstanding: 1) This misinterpretation of model cannot be easily recognized with classical cross-validation procedure when data are randomly split into training and testing sets because data samples of the same individual could appear in both sets. 

In this paper, we propose a Select-Additive Learning (SAL) procedure that addresses the confounding factor problem, specifically for neural architectures such as convolutional neural networks. 
Our proposed SAL approach is a two-phase procedure with the (\textit{Selection} phase and \textit{Addition} phase. During the \textit{Selection} phase, SAL identifies the confounding factors from the latent representation learned by neural networks. During the \textit{Addition} phase, SAL forces the original model to discard (or rather, give less importance to) the confounding elements by adding Gaussian noises to these representations. 
We conduct extensive experiments to test the performances of state-of-art neural-based models enhanced by SAL. All our experiments are performed in a person-independent setting, where subjects in the test set are different from the training and validation sets. We test the generalization with both, within-data and across-datasets experiments. 
%\footnote{The problem of confounding factors cannot be easily recognized with classical cross-validation procedure when data are randomly split into training and testing sets because data samples of the same person could appear in both sets. In person-independent setting, we constraint that the person can only appear in either training set or testing set.}. 

%The experiment results indicate that the SAL procedure works significantly better than the original model for each of the modalities separately as well as in the case of their fusion. 

% Our contribution of this papers are three fold:

% \textbf{$\star$} We validate the existence of ``individual's identity as confounding factor'' problem for multimodal sentiment analysis and show that state-of-the-art model does not well in real-world setting. 

% \textbf{$\star$} We propose a SAL procedure that can solve confounding factor problem.  

% \textbf{$\star$} We improve the prediction accuracy upon state-of-the-art significantly on each modality (audio, video, text), as well as on multimodal fusion, across data sets. 

%In the next section, we present a literature survey related to multimodal sentiment analysis. 
%Then we describe the SAL procedure which overcomes the issue of confounding factors, followed by a discussion of our experimental results and conclusions.

\section{Related Work}
\label{sec:related}

Multimodal data has been studied for a variety of applications to analyze human behaviors, including person detection and identification \cite{wu2016person,zhu2016distance}, human action recognition \cite{tejero2016human,zhao2016recognizing}, face recognition \cite{wu2016one,chen2016weakly}, as well as sentiment analysis. 

Originating from analysis of the textual modality, sentiment analysis has been carried out at the word level \cite{cambria2014senticnet}, phrase level \cite{wilson2005recognizing} and sentence level \cite{riloff2003learning}. % phrase level \cite{wilson2005recognizing}, sentences level \cite{riloff2003learning} and document level \cite{pang2004sentimental} extensively. %Recently, deep neural networks have also been used \cite{socher2013recursive}. 
\cite{kaushik2013sentiment} performed sentiment analysis on audio data by first transcribing the spoken words and then performing sentiment analysis. Related to audio-based sentiment analysis is the task of estimating emotional state of the speaker from audio input \cite{wuinferring}. 
For the visual modality, the Facial Action Coding System \cite{ekman1977facial} laid the groundwork for analyzing facial expressions and emotions. Recently, convolutional neural networks were used to discover the affective regions for sentiment on still images \cite{sun2016discovering}. 

The fusion of textual, acoustic and visual modalities for sentiment analysis has drawn increasing attention lately \cite{morency2011towards}. A variety of methods have been proposed and extensively discussed in recent years \cite{perez2013utterance,casaburi2015magic,poria2016fusing}. The state-of-the-art performance is achieved with a Convolutional Neural Network \cite{poria2015deep}. 

Our proposed Select-Additive Learning (SAL) procedure improves the generalizability of neural networks. Our experiments show improved prediction accuracy for all three modalities (verbal, acoustic and visual) as well as for multimodal fusion. 
%We have elaborated the first point in the introduction. 
The following section introduces our proposed Select-Additive Learning procedure. 

\section{Select-Additive Learning}
%\subsection{Definitions and Notations}
The main goal of our work is to increase the generalizability of multimodal sentiment prediction models by encouraging the model to consider sentiment-associated features (i.e. people are smiling while expressing positive sentiment) more than the identity-related features (i.e. wearing glasses). 
%Refer to Figure~\ref{fig:problem}(c), SAL aims to first identify the red dimensions (denoted as \textit{identity-related confounding dimensions}, as previously mentioned) and then force the model to ignore them. 
% To better illustrate SAL, we define the following terms:
% \begin{itemize}
% \item \textbf{General Representation}: The representation that is learned by neural networks from features associated with sentiment, in contrast to \textbf{Confounding Representation}
% \item \textbf{Confounding Representation}: The representation that is learned by neural networks from confounding factors, which is not associated with sentiment and is idiosyncratic.
% \item \textbf{Mixed Representation}: The representation that results from a combination of the \textbf{Causal Representation} and \textbf{Confounding Representation}.
%\end{itemize}
%We will use the notations summarized in Table~\ref{tab:notations}.

We formalize the problem by defining an input feature matrix $X$ of size $n \times p$ that encodes the $p$ features for $n$ utterances. In the multimodal scenario, p will be the total number of verbal, acoustic and visual features. We also define a vector $y$ of size $n\times 1$ which represents the sentiment of each utterance. Finally, we define a new matrix $Z$ which encodes for each utterance the speaker identity in a one-hot matrix of size $n \times m$ where $m$ represents the total number of unique individuals in the dataset.
%\footnote{for video data, it is trivial to bunch together videos of the same person to obtain identity information}.

\subsection{Select-Additive Learning Architecture}
\begin{figure}
\centering
  \includegraphics[width=0.35\textwidth]{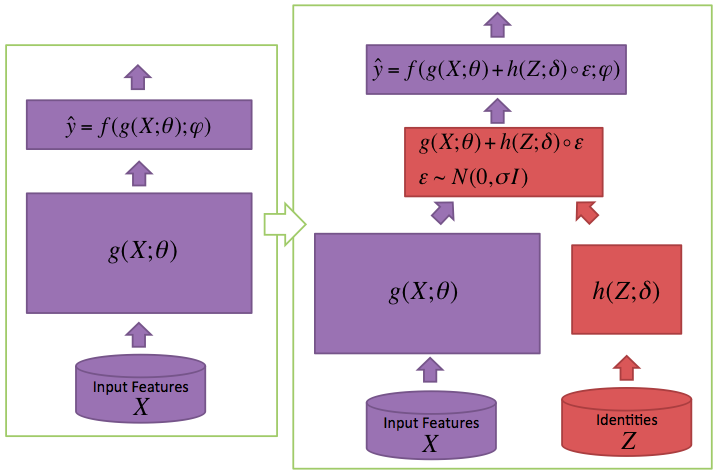}
  \caption{The SAL architecture is achieved by a simple extension of a general deep learning discriminative classifier. The purple part is the original deep learning model. %that could be split into representation learner $g(\cdot; \theta)$ and  discriminative classifier $f(\cdot, \phi)$. 
The red part is the extension SAL introduces. 
%$h(\cdot; \delta)$ stands for a simple neural network. $\theta$, $\phi$ and $\delta$ stand for the parameters of each network. 
The extension network is connected to the original network via a Gaussian Sampling Layer.}
  \label{fig:model}
\end{figure}
Our proposed SAL procedure is designed to enhance a pre-existing (i.e. pre-trained model) discriminative neural network to be more robust against confounding factors. 
To formally introduce our SAL approach, we define two main components present in most discriminative neural network classifiers (e.g., Convolutional Neural Network, CNN): a representation learner component and a classification component. 
To simplify the notation, we use $g(\cdot; \theta)$ denotes the representation learner component and $\theta$ stands for its parameters. Our hypothesis is that confounding factors will be constrained to a subset of dimensions present in $g(\cdot; \theta)$. Similarly, we use $f(\cdot; \phi)$ to denote the classification component and $\phi$ denotes the parameters. Therefore, a full neural network classifier is denoted as $f(g(\cdot; \theta);\phi)$.
In our SAL approach, $g(\cdot; \theta)$ from identity-related features as \textit{identity related confounding dimensions}. Our SAL approach can be summarized as first identifying these dimensions and then reduce the impact of these dimensions by adding noise to them.

%For example, in a typical CNN, the representation learner component consists of convolutional layers, pooling layers and MLP, while the classification component is a Logistic Regression Layer. 

To select \textit{identity-related confounding dimensions}, SAL introduces a simple neural network (denoted by $h(\cdot; \delta)$, where $\delta$ stands for its parameters). This is to predict \textit{identity-related confounding dimensions} from individual identities $Z$, by minimizing the difference between $h(Z; \delta)$ and $g(X; \theta)$. Therefore, $h(Z; \delta)$ will effectively pinpoint the \textit{identity-related confounding dimensions} in $g(X; \delta)$. Figure 3a shows an overview of this Selection Phase.

To force the model to discard \textit{identity-related confounding dimensions}, SAL introduces Gaussian noise to these dimensions while minimizing prediction error, so that $f(\cdot; \phi)$ learns to neglect noised representation. The noise is added through a Gaussian Sampling Layer \cite{kingma2013auto}. Figure 3b shows an overview of this addition phase.

Figure~\ref{fig:model} shows how SAL assembles $g(\cdot; \theta)$, $f(\cdot; \phi)$ and $h(\cdot; \delta)$ together via a Gaussian Sampling Layer.

\subsection{Select-Additive Learning Algorithm}

A pre-requisite to our Select-Additive Learning (SAL) approach is first learn a discriminative neural classifier. On our experiments, we achieve this goal by minimizing the following lost function:
\begin{align*}
\arg\!\min_{\phi, \theta} \dfrac{1}{2}(y - f(g(X; \theta); \phi))^2
%\label{eq:prior}
\end{align*}
The same loss function is often used in discriminative neural networks \cite{wang2017origin}.

\begin{figure}[ht]
\centering
\subfloat[\textit{Selection} Phase: SAL forces $h(Z; \delta)$ to identify \textit{identity-related confounding dimensions}. Right side figure shows that $h(Z; \delta)$ selects these dimensions. ]{
  \includegraphics[clip,width=0.8\columnwidth]{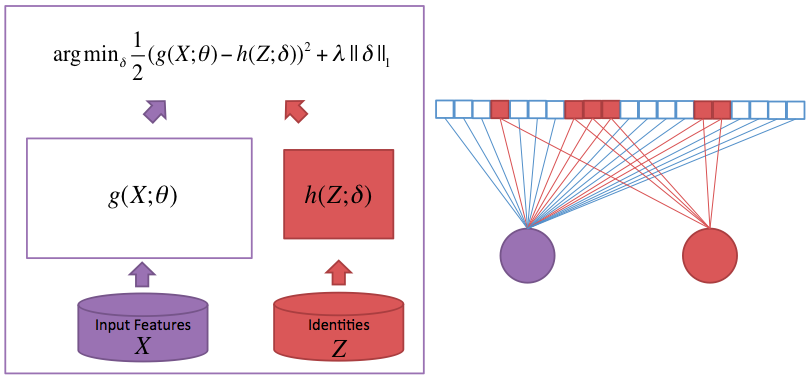}
}\\
\subfloat[\textit{Addition} Phase: SAL forces the model to focus on other dimensions by adding Gaussian noise to \textit{identity-related confounding dimensions}. Right side shows that the model shifts focus because these dimensions are contaminated/noisy.]{
  \includegraphics[clip,width=0.8\columnwidth]{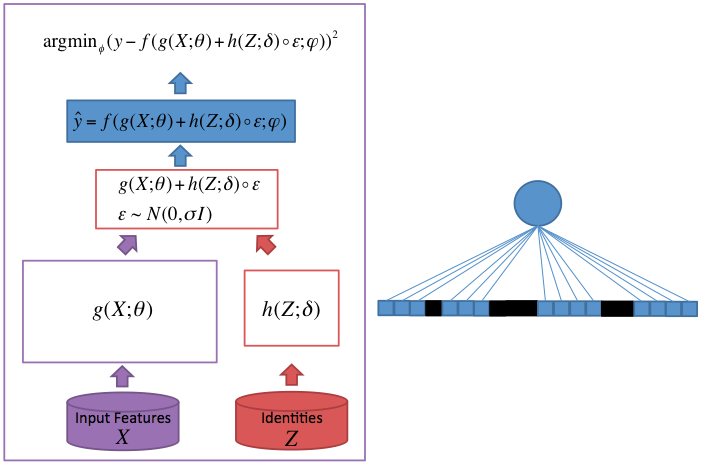}
}
\caption{Illustration of SAL. On the left, network structure and training objective is presented. On the right, circles denote neurons. Squares denote dimensions of representation. 
%Blue, red and purple follow the coloring convention of this paper. 
%Black stands for contaminated dimensions.
}
\label{fig:algo}
\end{figure}

\subsubsection{\textit{Selection} Phase}
Once the original representation g(X;$\theta$) is learned, the selection phase optimizes a new loss function to discover the identity related confounding dimensions. This selection phase is operationalized by tuning the parameters $\lambda$ using the following loss function (as illustrated in Figure~\ref{fig:algo} (a)):
\begin{align}
\arg\min_{\delta}\dfrac{1}{2}(g(X; \theta)-h(Z; \delta))^2 + \lambda ||\delta||_1
\label{eq:sel}
\end{align}
where $\lambda$ is a scalar that controls the weight of the sparsity regularizer. 
In this phase, both $X$ and $Z$ are available, but only $\delta$ is tuned, as shown in Fig.~\ref{fig:algo}(a). 

The goal of this phase is to select \textit{identity-related confounding dimensions} from the original representation. To achieve this, we tune $\delta$ to minimize the difference between $g(X;\theta)$ and $h(Z; \delta)$. As $Z$ only encodes identity information, the minimum of difference will be achieved when $h(Z; \delta)$ is matched to the \textit{identity-related confounding dimensions} of $g(X; \theta)$. L1 regularization of $\delta$ is necessary to avoid overfitting as output dimension of $h(\cdot; \delta)$ is typically significantly higher than input dimension.

The result of this \textit{selection} phase is shown on the right-hand-side of Fig.~\ref{fig:algo}(a). All the weights of original model (purple circle) are active and connected to every dimension while only some weights of $h(\cdot, \delta)$ (red circle) are active and connected to the \textit{identity-related confounding dimensions}

\subsubsection{\textit{Addition} Phase}
After the selection phase $h(\cdot, \delta)$ should be pointing at the \textit{identity-related confounding dimensions}. Our remaining step is to learn a new neural network classifier where the confounding dimensions have "masked". We achieve this by adding Gaussian noise.
Our addition phase defines the following loss function to achieve this goal:
%and finally our objective function to be minimized becomes \aaksha{Number the equation mentioned earlier and just mention the no. here}:
\begin{align}
\arg\!\min_{\phi} \dfrac{1}{2}(y - f(g(X; \theta)+h(Z; \delta)\circ\epsilon; \phi))^2
\label{eq:add}
\end{align}
where $\epsilon \sim N(0, \sigma I)$ and $\circ$ stands for element-wise product, , as showed in Fig.~\ref{fig:algo}(b). 

In this phase, parameter $\phi$ is tuned. The input representation of $f(\cdot; \phi)$ consists of the representation learned from $g(X; \theta)$ and the $h(Z; \delta)$-selected \textit{identity-related confounding dimensions} with Gaussian noise added.
The noise ensures that \textit{identity-related confounding dimensions} are no longer informative so that $f(\cdot; \phi)$ can be trained to ignore them.

As illustrated on the right side of Fig.~\ref{fig:algo}(b), \textit{identity-related confounding dimensions} are contaminated with addition of noise. 
Therefore, the model learns to discard these non-informative dimensions, and its weights get optimized to focus on the rest of the dimensions.

%SAL introduces one more set of parameters to train ($\delta$) and two more hyper-parameters to select ($\lambda$ in Equation~\ref{eq:sel} and $\sigma$ in Equation~\ref{eq:add} via $\epsilon$). 
%Strategies to select $\lambda$ and $\sigma$ to tune $\delta$ are discussed in the supplementary material. 

\section{Experiments}
\label{sec:exp}
In this section, we perform an extensive set of experiments on three different data sets to see whether SAL can help improve the generalizability of a discriminative neural classifier. Generalizability is tested by performing across-dataset experiments where two of the dataset are kept exclusively for testing. All our experiments follow a person independent methodology where none of the subject from the training data are  present in the test datasets.
% \begin{table*}
% \centering
% \caption{Test accuracy on three data sets for CNN and SAL-CNN over three modalities and multimodal fusion. The models are trained and selected on the data of 62 individuals of MOSI data set, then tested on the rest 31 individuals. The same model is also tested across data set in YouTube data set and MOUD data set.}
% \label{tab:result}
% \begin{tabular}{cc|cc|cccc}
% \hline
% \multicolumn{1}{l}{} & \multicolumn{1}{l|}{} & \multicolumn{2}{c|}{Within Dataset} & \multicolumn{4}{c}{Across Dataset} \\
%  &  & \multicolumn{2}{c|}{MOSI} & \multicolumn{2}{c}{YouTube} & \multicolumn{2}{c}{MOUD} \\
%  &  & CNN & SAL-CNN & CNN & SAL-CNN & CNN & SAL-CNN \\ \hline
%  & Text & 0.678 & \textbf{0.732} & 0.605 & \textbf{0.657} & 0.522 & \textbf{0.569} \\
% Single Modality & Audio & 0.588 & \textbf{0.618} & 0.441 & \textbf{0.564} & 0.455 & \textbf{0.549} \\
%  & Video & 0.572 & \textbf{0.636} & 0.492 & \textbf{0.549} & \textbf{0.555} & 0.548 \\
%  & Text+Audio & 0.687 & \textbf{0.725} & 0.642 & \textbf{0.652} & 0.515 & \textbf{0.574} \\
% Double Modalities & Text+Video & 0.706 & \textbf{0.73} & 0.642 & \textbf{0.667} & 0.542 & \textbf{0.574} \\
%  & Audio+Video & \textbf{0.661} & 0.621 & 0.452 & \textbf{0.559} & 0.533 & \textbf{0.554} \\
% \multicolumn{2}{c|}{All Modalities} & 0.715 & \textbf{0.73} & 0.611 & \textbf{0.667} & 0.531 & \textbf{0.574} \\ \hline
% \end{tabular}
% \end{table*}

\subsection{Models}
\label{sec:model}
We compare the following models:

\textbf{$\star$CNN}: The state-of-the-art seven layer convolutional neural network architecture used previously for multimodal sentiment analysis \cite{poria2015deep}. 

\textbf{$\star$SAL-CNN}: After the state-of-the-art CNN is fully trained, we use SAL to increase its generalizability and predict sentiment. $h(\cdot, \delta)$ is a neural perceptron \cite{wang2017origin}. 

\subsection{Datasets}
\label{sec:data}
We performed our experiment on three multimodal sentiment analysis data sets: 

\textbf{$\star$MOSI}: This dataset consists of 93 videos obtained from YouTube channels. Each video contains the opinions from one unique individual. The dataset has 2199 utterances manually segmented from online videos of movie reviews. Each utterance was also manually annotated for sentiment label \cite{zadeh2015micro}.

\textbf{$\star$YouTube}: This dataset consists of 47 opinion videos with 280 utterances with manually annotated sentiment labels \cite{morency2011towards}. Each video contains the opinions from one unique individual.

\textbf{$\star$MOUD}: This dataset consists of 498 Spanish opinion utterances from 55 unique individuals \cite{rosas2013multimodal}.

Although, majority of the data originate from YouTube, they differ in recording quality and the processing done after curation. 
%In addition, verbal features are obtained from different ASR tools for all the three datasets. 
The verbal features in the MOUD dataset need one extra step of translation from Spanish to English. These three datasets are good candidates to evaluate across-dataset generalization.

\subsection{Feature Extraction}
We extracted an embedding for each word using a word2vec dictionary pre-trained on a Google News corpus \cite{mikolov2013efficient}. 
%We also appended a 6 dimensional binary vector to this embedding to indicate the POS tag of the word (noun,verb,adjective,adverb, preposition, conjunction). 
The text feature of each utterance was formed by concatenating the word embeddings for all the words in the sentence and padding them with the appropriate zeros to have the same dimension. We set the maximum length as 60 and discarded additional words\footnote{only around $0.5\%$ utterances in our datasets have more than 60 words}. 
%The penultimate fully connected layer of the CNN is extracted to form the final text input for training. 
For YouTube dataset, we extracted the transcripts using the IBM Bluemix's speech2text API\footnote{https://www.ibm.com/watson/developercloud/speech-to-text.html}. For MOUD dataset, we translated Spanish transcripts into English transcripts. 
We used openSMILE \cite{eyben2010opensmile} to extract the low-level audio descriptors for each spoken utterance. These audio descriptors included the Mel-frequency cepstral coefficients, pitch and voice quality. %We split each utterance into 50 trunks and took the average of features within each trunk and resulted in a set of 1950 dimensional vectors for each utterance. 
We processed every frame in each video and used the audio-visual synchrony to identify which frames happen during a specific utterance. We used the CLM-Z library \cite{baltruvsaitis20123d} for extracting facial characteristic points. %We split each utterance into 5 trunks and took the average of features within each trunk and resulted in a set of 2075 dimensional vectors for each utterance. 

\subsection{Experiment Setup}
\label{sec:exp_set}
%We simulated a real-world setting by applying the constraint that training and validation set had no individuals in common with the test set. 
We remove the netural utterances out of the data set. 
The first 62 individuals in the MOSI data set are selected as training/validation set. There are around 1250 utterances in total. These utterances are shuffled and then 80\% are used for training and 20\% used for validation. We have three test datasets.
1) MOSI: 546 utterances from the remaining 31 individuals. 2) YouTube: 195 utterances from 47 individuals and 3) MOUD: 450 utterances from 55 individuals. We use MOSI as training set because it is the largest and most recent dataset among all three.

\subsection{Experiment Results}
%To make a more concrete comparison, we split the experiments to compare models within the same distribution of data and across distributions of data. 
\subsubsection{Within data set}
\begin{table}[]
\centering
\caption{Within data set experiments}
\label{tab:result1}
\begin{tabular}{cccc}
\hline
 &  & \textbf{CNN} & \textbf{SAL-CNN} \\ \hline
\multirow{3}{*}{Unimodal} & Verbal & 0.678 & \textbf{0.732} \\
 & Acoustic & 0.588 & \textbf{0.618} \\
 & Visual & 0.572 & \textbf{0.636} \\ \hline
\multirow{3}{*}{Bimodal} & Verbal+Acoustic & 0.687 & \textbf{0.725} \\
 & Verbal+Visual & 0.706 & \textbf{0.73} \\
 & Acoustic+Visual & \textbf{0.661} & 0.621 \\ \hline
\multicolumn{2}{c}{All Modalities} & 0.715 & \textbf{0.73} \\ \hline
\end{tabular}
\end{table}

Table~\ref{tab:result1} shows the results for \textbf{CNN} and \textbf{SAL-CNN} tested on the remaining 31 individuals' data of MOSI. The results 
%of \textbf{SAL-CNN} are an improvement by about 5\% on average, which 
indicate that SAL could help to increase the generalizability of the trained model. 

\subsubsection{Across data sets}
\begin{table}[]
\centering
\caption{Across data set experiments}
\label{tab:result2}
\begin{tabular}{ccccc}
\hline
 & \multicolumn{2}{c}{Youtube} & \multicolumn{2}{c}{MOUD} \\
 & \textbf{CNN} & \textbf{SAL-CNN} & \textbf{CNN} & \textbf{SAL-CNN} \\ \hline
Verbal & 0.605 & \textbf{0.657} & 0.522 & \textbf{0.569} \\
Acoustic & 0.441 & \textbf{0.564} & 0.455 & \textbf{0.549} \\
Visual & 0.492 & \textbf{0.549} & \textbf{0.555} & 0.548 \\ \hline
Ver+Acou & 0.642 & \textbf{0.652} & 0.515 & \textbf{0.574} \\
Ver+Vis & 0.642 & \textbf{0.667} & 0.542 & \textbf{0.574} \\
Acou+Vis & 0.452 & \textbf{0.559} & 0.533 & \textbf{0.554} \\ \hline
All & 0.611 & \textbf{0.667} & 0.531 & \textbf{0.574} \\ \hline
\end{tabular}
\end{table}

Table~\ref{tab:result2} shows the results for \textbf{CNN} and \textbf{SAL-CNN} tested on YouTube and MOUD dataset. First, it is noteworthy that in some cases the performance of the CNN is worse than mere chance. This inferior performance substantiates the existence of the non-generalization problems we are targeting. %because these models are selected as the ones that achieve minimum error rate in validation sets and they can barely perform well when tested across data sets.

%Text modality performs the best among the three modalities. We conjecture two reasons for this: 1) Sentiment is better defined over textual than over visual or audio features, and textual information may strongly affect label decisions during annotation. 2) Language is more informative than visual or audio expressions, and less dependent on individuals' identities. However, our results show that, even on text modality, the learning process of models may still be confounded as individuals may have different preferences of words and sentence structure. This is especially true when considering that our textual data are collected from spoken languages. 

%All the three data sets come from the same web platform, they may have been through different processing procedures \haohan{to be verified}, which results in slightly different distributions and leads to that testing accuracies on YouTube and MOUD are much lower than that of MOSI in audio modality and video modality.
%These differences show up in the accuracies for the YouTube and MOUD datasets which are much lower than those of MOSI.
%These differences are also reflected in accuracies for the text modality. 
Overall, Select-Additive Learning increases the robustness and performance of the previous models consistently (except only two cases: Video modality in MOUD and fusion of acoustic \& visual in MOSI). Permutation Test rejects the null hypothesis (no improvement) with p-values $0.037$, $0.0003$, $0.0023$ respectively for MOSI, YouTube, and MOUD, indicating significant improvement.\footnote{Select-additive Learning implementation is available at \\https://github.com/HaohanWang/SelectAdditiveLearning}  

\subsection{Discussion}

\begin{figure}[]
  \centering
  \includegraphics[width=0.8\columnwidth]{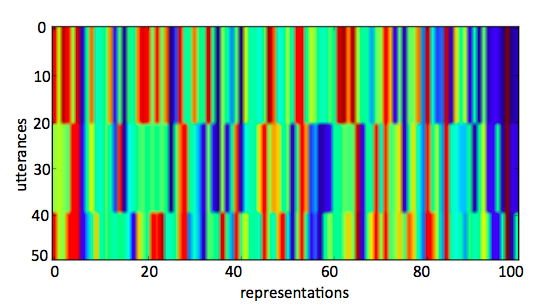}
  \caption{Confounding factors identified in the \textit{Selection} phase for first 50 utterances (rows), first 100 representation values (columns) in the training set. %This figure shows that these first 50 utterances clearly belong to three different individuals, increasing the chance of confounding factor between sentiment and identity.
  }
  \label{fig:conf}
\end{figure}

To substantiate our proposed model and algorithm, we examine the learning process and verify that representation of confounding factors exists and our method can mitigate its effects. We demonstrate this with the visual modality as it intuitively contributes the most to confounding. 

Figure~\ref{fig:conf} shows a plot of $h(Z, \delta)$ during the \textit{Selection} phase. It is a zoomed-in figure for the first 50 utterances (rows) and first 100 values of the representation vector (columns). Blue indicates lowest values and red indicates highest values and other colors are linearly interpolated. 

The representation of utterances forms clear clusters and each cluster belongs to one person. %This figure suggests that confounding representation might be different across individuals, which fosters the argument in favor of \textit{Addition} phase of Select-Additive Learning as opposed to dropping weights. 
Despite each individual having their own pattern, some dimensions have generalized well across individuals. Our model learns to assign more weights to these dimensions after noise is introduced. 

In addition to these results, we calculated the inter-cluster distance over intra-cluster distance ratio for the representation learned under two situations: 1) clustered by category of sentiment and 2) clustered by individual's identity. We compared the ratios for \textbf{CNN} and \textbf{SAL-CNN}. The higher ratio indicates a clearer clustering structure. 

After SAL, for representation clustered by category of sentiment, the ratio increased by 44\%, 15\% and 72\% respectively for verbal, acoustic, and visual modality, while for representation clustered by individual's identity, the ratio increased by 9\%, 3\% and 13\%, respectively. These numbers indicate SAL almost maintains the clustering structure of identity, but greatly improves the clustering structure of category of sentiment. This shows the effectiveness of SAL. 

%These results are discussed in detail in supplementary materials. 

\section{Conclusion}
High-quality datasets required to train machine learning models for automatic multimodal sentiment analysis are only of the order of a few thousand samples. These limited resources restrict models' generalizability, leading to the issue of confounding factors. 
We proposed a Select-Additive Learning (SAL) procedure that can mitigate this problem. 
With extensive experiments, we have shown how SAL improves the generalizability of state-of-the-art models. We increased prediction accuracy significantly in all three modalities (verbal, acoustic, visual), as well as in their fusion. We also showed that SAL could achieve good prediction accuracy even when tested across data sets. 
%In this paper, we first presented the existence of a problem for multimodal sentiment analysis. The sentiment is not independent of individuals and a model could be confounded by individual-sensitive features such as appearance. Therefore, a model trained on a group of individuals does not generalize well to other individuals.

%After verifying the existence of the problem, we proposed a Select-Additive Learning procedure to solve it. 
%SAL is a two-phase learning method. In \textit{Selection} phase, it selects the \textit{identity-related confounding dimensions}. In \textit{Addition} phase, it forces the classifier to discard these dimensions by adding Gaussian noise. 
%In our experiments, we showed how SAL improves the generalizability of state-of-the-art models. We increased prediction accuracy significantly in all three modalities (text, audio, video), as well as in their fusion. We also showed that SAL could achieve good prediction accuracy even when tested across data sets. 

\bibliographystyle{IEEEbib}
\bibliography{icme2017template}

\end{document}